\documentclass{Interspeech}

\interspeechcameraready 

\title{Voice Activity Projection Model with Multimodal Encoders}

\author[affiliation={1}]{Takeshi}{Saga}
\author[affiliation={1,2}]{Catherine}{Pelachaud}

\affiliation{ISIR}{Sorbonne University}{France}
\affiliation{}{CNRS}{France}
\email{takeshi.saga@sorbonne-universite.fr, catherine.pelachaud@sorbonne-universite.fr}

\usepackage{comment}

\begin{document}

    \maketitle
    
    \begin{abstract}
        %
        %
        Turn-taking management is crucial for any social interaction. 
        Still, it is challenging to model human-machine interaction due to the complexity of the social context and its multimodal nature. 
        Unlike conventional systems based on silence duration, previous existing voice activity projection (VAP) models successfully utilized a unified representation of turn-taking behaviors as prediction targets, which improved turn-taking prediction performance. 
        Recently, a multimodal VAP model outperformed the previous state-of-the-art model by a significant margin. 
        In this paper, we propose a multimodal model enhanced with pre-trained audio and face encoders to improve performance by capturing subtle expressions. 
        Our model performed competitively, and in some cases, even better than state-of-the-art models on turn-taking metrics.
        All the source codes and pretrained models are available at \url{https://github.com/sagatake/VAPwithAudioFaceEncoders}.
    \end{abstract}

    \section{Introduction}




        
        Turn-taking is a fundamental and universal function in social interactions, enabling the coordinated exchange of speaking turns between participants in a conversation, which is essential for successful social communication \cite{Sacks1974,O-Connell1990}. 
        Smooth turn-taking enables coherent dialogue, preventing everyone from speaking simultaneously. 
        We can divide turn-taking cues into three aspects: turn-yielding cues, turn-holding cues, and turn-initial cues \cite{Duncan1972}. 
        Turn-yielding cues (e.g., the falling pitch at the end of the sentence) indicate the speaker's intentions to yield. 
        In contrast, turn-holding cues (e.g., level intonation) show that the speaker wants to continue their turn. 
        Unlike these cues, turn-initial cues (e.g., raising an index finger) can be produced by the following speaker candidates to explicitly indicate that they wish to speak next or, in some cases, even interrupt the current speaker's speech. 
        These cues are multimodal.
        In particular, the cues include text (e.g., syntax, pragmatics, semantics, fillers) \cite{Ford1996}, speech (e.g., pitch, intensity, formants) \cite{Ward2019}, pauses (filled pause, unfilled pause) \cite{Clark2002}, hand/arm gestures (termination, tense, rhythmic) \cite{Duncan1972}, head movement (nodding, shaking, tilting) \cite{Ward1996}, postures (leaning, swelling) \cite{Padilha2003}, and so on. 
        Interlocutors produce and rely on these cues to take, yield, or hold turns, where reproduction is challenging in robots or embodied conversational agents.
        
        In addition to this multimodal understanding, turn-taking timing is another crucial factor for smooth dialogue, especially in sustained interactions between humans and robots or humans and agents \cite{Robins2006}. 
        Participants perceived agents with too early responses (interruptions) as less agreeable and more assertive \cite{Cafaro2016, Yang2023}. 
        In contrast, agents with longer pauses give the impression of creating more rapport \cite{ter-Maat2010, Cafaro2016}.
        Although human language production systems require about 600 ms to produce the first response word, the average response time from the end of the preceding utterance is about 200 ms \cite{Levinson2015}. 
        This implies that humans may use a predictive approach to content or timing. 
        As Meyer mentioned, sentence processing in the human brain is highly incremental and predictive, with evidence that speakers can prepare utterances while listening to another person's speech \cite{Meyer2023}. 
        For example, speakers can rapidly grasp the content and speech acts. 
        They can also predict the ends of turns quickly and precisely. 
        Several turn-taking management models have been proposed to simulate this incremental and predictive sentence processing \cite{DeVault2009, Hough2016}.

        Turn-taking management is vital for sustained interaction between humans and artificial agents \cite{Robins2006}. 
        Participants perceived agents with too early responses (interruptions) as less agreeable and more assertive \cite{Cafaro2016, Yang2023}. 
        In contrast, agents with longer pauses give the impression of creating more rapport \cite{ter-Maat2010}.
        ERICA by Lala et al. is an iconic example of automatic human-robot turn-taking \cite{Lala2017}.
        The authors used logistic regression models separately for predicting backchannel generation and turn-taking timing. 
        They calculate prosodic features, such as pitch and intensity, over multiple time windows and predict the likelihood of occurrence within the next 500 milliseconds. 
        Instead of predicting backchannels and turn-taking timing separately, Ekstedt proposed to predict them with a unified representation called voice activity projection state \cite{Ekstedt2022-a}. 
        Using the unified representation, they successfully predicted both the timings of the turn shift and the back channel. 
        However, this model relies only on audio signals. 
        It cannot predict turn-taking timings that involve cues such as gaze direction, mouth opening, and other nonverbal gestures \cite{Ishii2013, Tian2018, Ishii2019}.
        %
        %
        Recently, Onishi et al. proposed a model that takes as input nonverbal signals (facial action units corresponding to facial muscular contractions, head poses, and body joint positions) and audio signals \cite{Onishi2023}. 
        Although the model improved prediction performance, these nonverbal features can represent only superficial characteristics, excluding context-rich information such as sequential, spatial, relational, and temporal relationships between them. 
        Other researchers have proposed text-based LLM approaches, but they still have challenges running in an auto-regressive manner in nearly real-time \cite{Wang2024, Pinto2024}. 
        More recently, though speech-to-speech foundation models such as Mini-Omni or Moshi (Gemini Live or ChatGPT voice chat as service-level examples) showed very natural speech conversation abilities, investigations on internal mechanisms of those fully end-to-end models are still ongoing research, which is generally very challenging in terms of explainability \cite{Xie2024, Defossez2024, GeminiLive, ChatGPTVoiceChat}.
        
        In this paper, we aim to thoroughly consider nonverbal turn-taking cues, proposing to enhance Ekstedt's voice activity projection model with pre-trained encoders for audio and facial signals \cite{Ekstedt2022-a}. 
        We hypothesize that combining the encoders will contribute to capturing the coordinated, multimodal intentions of the speakers, conveyed by both facial and audio signals. 
        Specifically, by adding a facial image encoder, the subtle social context of facial expressions will be embedded into the network. 
        We trained the multimodal prediction models to prove this idea and evaluated their performance. 
        In the remainder of this paper, after presenting the variants of VAP models, we describe our model.

    \section{Voice activity projection}

\begin{figure}
            \centering
            \includegraphics[width=1.0\linewidth]{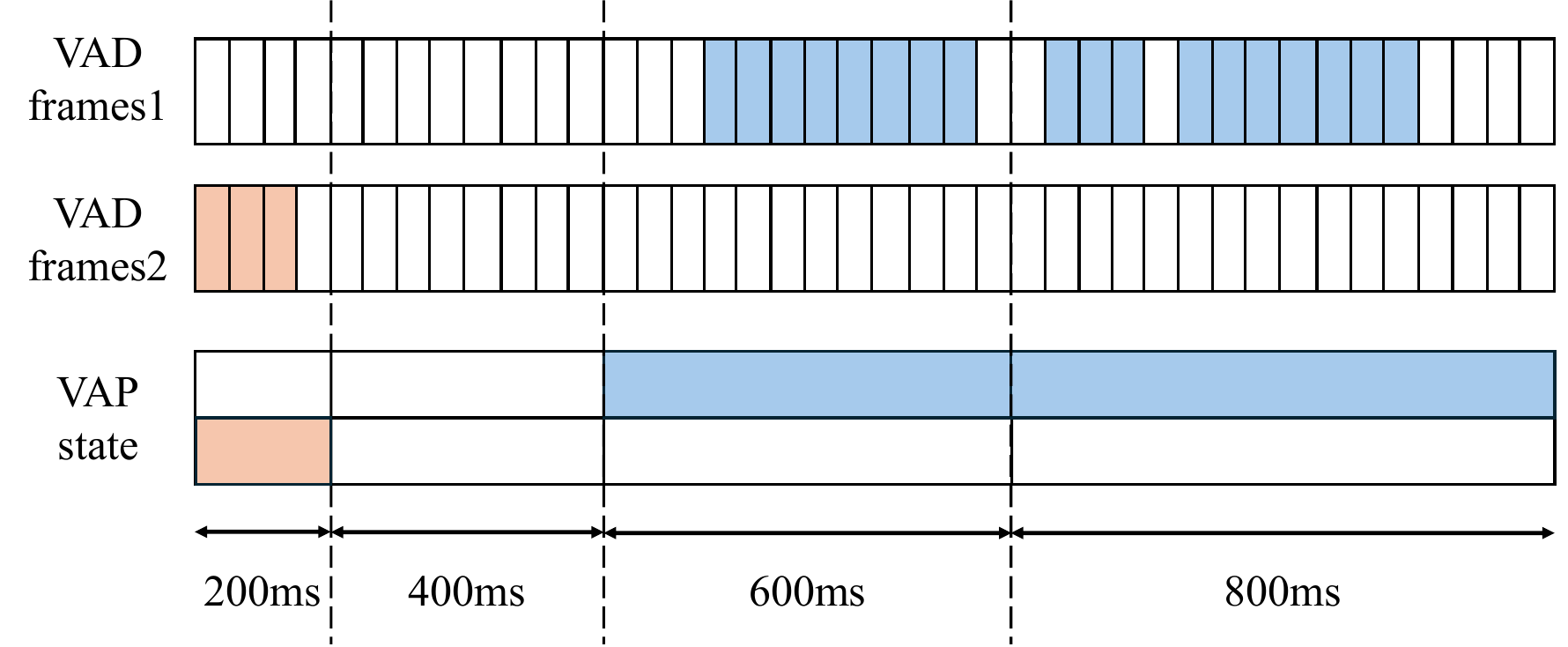}
            \caption{VAP window}
            \label{fig:vap_window}
        \end{figure}

        One of the successful models in turn-taking prediction is the voice activity projection (VAP) by Ekstedt et al. \cite{Ekstedt2022-a}. 
        Unlike conventional voice activity detection (VAD) models, which consider only the voice activity of a single speaker, the VAP model predicts both users' speaking states (e.g., speaking or not speaking) and unified VAP states, aiming to model interactions better for turn-taking behavior prediction. 
        Additionally, the VAP model employs a predictive approach, predicting what will happen next based on the previous audio signals.
        %
        Figure~\ref{fig:vap_window} shows an example of the VAP state. 
        VAP state is a discrete representation shaped as 2 × 4 binary bins. 
        Each row corresponds to a time series of voice activity of each user. 
        Based on the assumption that further activities are more challenging to predict, the duration of each time bin is defined as 0.2, 0.4, 0.6, and 0.8 sec. from the nearer future to the further future. 
        In the training period, the model is trained by multi-task learning of future VAD prediction for each user and VAP state.
        This model has been validated in various aspects, including real-time prediction systems, the training effect on a multilingual dataset, and the prosody effect on synthesized voice by text-to-speech \cite{Inoue2024-a, Inoue2024-b, Ekstedt2022-b}.
        
        Onishi et al. extended this VAP model by incorporating nonverbal signals, including gaze angles, facial action units, head poses, and upper body joint coordinates \cite{Onishi2023}. 
        Those features were extracted using OpenFace2.0 (gaze, face, head) and OpenPose (body) \cite{Baltrusaitis2016, Cao2021}.
        While the authors reported that the nonverbal features significantly improved the prediction performance, the facial action units contributed the most to the prediction of turn-taking behavior. 
        However, there remains room for improvement since the facial action unit only captures superficial facial expressions for each image frame and does not include sequential dynamics and subtle expressions, which are difficult to code explicitly. 

    \section{Proposed model}

        \begin{figure}
            \centering
            \includegraphics[width=1.0\linewidth]{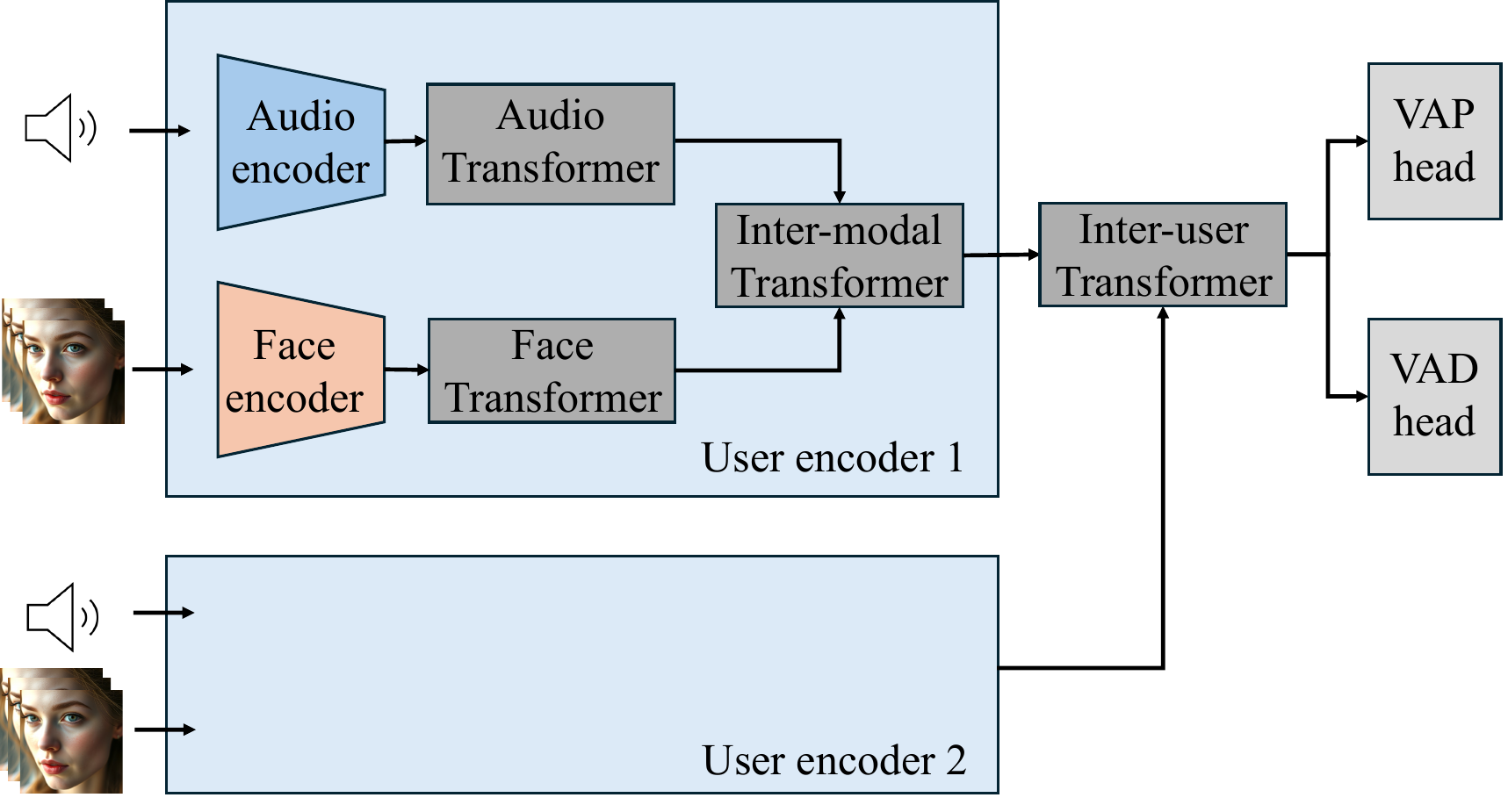}
            \caption{Architecture of proposed model including encoders}
            \label{fig:proposed_model}
        \end{figure}
        
        Figure~\ref{fig:proposed_model} shows the network architecture of our proposed multimodal model. 
        Unlike Onishi's multimodal model, we inserted a pretrained encoder for facial images instead of action units. 
        Since Onishi and colleagues reported that face information contributed the most, we hypothesized that richer representations extracted with a pre-trained encoder would improve performance. 
        In our work, we have searched the encoder from publicly available pretrained models and selected Dynamic Facial Expression Recognition Transformer (Former-DFER), which has been trained on the Dynamic Facial Expression in the Wild (DFEW) dataset for facial emotion recognition \cite{Zhao2021}. 
        This model achieved promising results in the social task (emotion recognition) and was further trained for video sequences. 
        We used the same pretrained Contrastive Predictive Coding (CPC) model as in Skantze and Onishi's works for the audio modality \cite{Riviere2020}. 
        We trained a model with the following input variants. 
        \begin{enumerate}
            \item \textit{Proposed1}: audio signal, face image sequence
            \item \textit{Proposed2}: audio signal, face image sequence, head angles, gaze angles, and normalized 2D body positions
            \item \textit{Proposed3}: audio signal, face image sequence, face AU sequence, head angles, gaze angles, and normalized 2D body positions
        \end{enumerate}
        Furthermore, we also changed our fusion methods. 
        Onishi's approach involved merging user signals for each modality separately and then fusing across different modalities.  We modified the architecture to capture interactions between modalities for each participant more effectively. 
        We first merged multimodal signals for each person separately, followed by the fusion across user embeddings. 
        
        Regarding input data, we set the processing frame rate to 25, following Onishi's work to align with NoXi's video frame rate. 
        The audio was recorded as stereo, with each speaker's voice in the right and left channels separately. 
        We used a sampling rate of 16 kHz, following Onishi and Ekstedt. 
        For image modality, we used extracted facial images as input. 
        We used the pretrained face detection model based on the max-margin convolutional neural network bundled with the dlib machine learning library \cite{King2009}. 
        First, we localized the face position in the original image. 
        Then, we clipped the face image based on the detected face rectangle and resized it to (3, 112, 112) to match the face encoder's input size. 
        As in Ekstedt and Onishi's paper, the input sequence length was 20 seconds, and the prediction window was 2 seconds.

        %
        %
        \begin{table*}[ht]
            \centering
            \begin{tabular}{||l||c|c|c|c||l|l|l|l||}
                \toprule
                Model           & audio     & face AU   & face embed& head, gaze, body  & S/H   & S/L   & SP    & BC \\
                \midrule
                \midrule
                \textit{Baseline1}       &\checkmark &           &           &                   & 0.712 & 0.684 & 0.726 & 0.434 \\ 
                \midrule
                \midrule
                \textit{Baseline2\_1}    &\checkmark &\checkmark &           &                   & \textbf{0.758} & 0.634 & 0.788 & 0.442 \\ 
                \midrule
                \textit{Baseline2\_2}    &\checkmark &\checkmark &           &\checkmark         & 0.735 & \textbf{0.716} & 0.768 & 0.428 \\ 
                \midrule
                \midrule
                \textit{Proposed1}           &\checkmark &           &\checkmark &                   & 0.703 & 0.625 & 0.709 & \textbf{0.506} \\ 
                \midrule
                \textit{Proposed2}           &\checkmark &           &\checkmark &\checkmark         & 0.687 & 0.703 & 0.730 & 0.446 \\ 
                \midrule
                \textit{Proposed3}           &\checkmark &\checkmark &\checkmark &\checkmark         & 0.737 & 0.628 & \textbf{0.794} & 0.503 \\ 
                

                
                \bottomrule
            \end{tabular}
            \caption{Turn-taking Behavior Prediction Accuracy for different conditions: \textit{Baseline1} involves only audio; 2 conditions of \textit{Baseline2}: \textit{Baseline2\_1} involves audio, face(AU); \textit{Baseline2\_2} involves audio, face(AU), head, gaze, body; 3 conditions for our model \textit{Proposed1} involves audio, face(encoder);  \textit{Proposed2} involves audio, face(encoder), head, gaze, body;  \textit{Proposed3} involves audio, face(AU), face(encoder), head, gaze, body.}
            \label{tab:performance}
        \end{table*}


    \section{Method}

        \subsection{Dataset: NoXi}

            We used the NoXi multimodal dataset for our experiments \cite{Cafaro2017}.
            NoXi is the dataset of screen-mediated multimodal face-to-face interactions.
            This dataset includes dyadic interaction videos from Germany, the United Kingdom, and France, and was recently extended with a Japanese subset \cite{Onishi2023}. 
            We used the French subset to simplify the setting, which resulted in 7 hours of videos and 21 interactions.

            Each participant was in a different room and interacted through a video conference format. 
            Although video and audio were recorded separately for each room, they are all synchronized with very low latency via wired LAN cables. 
            We can extract synchronized nonverbal behaviors, including body and head poses, arm gestures, and facial expressions, from each video. 

        \subsection{Experiment setting}


            To investigate the feasibility of our approach, we trained and compared our model with the Ekstedt model (\textit{Baseline1}) and Onishi's model (\textit{Baseline2}) \cite{Ekstedt2022-a, Onishi2023} under different conditions (see Table \ref{tab:performance}).
            We implemented the source code with the PyTorch Lightning framework \cite{pytorch_lightning}. 
            We extended Inoue's publicly available code implementation (\url{https://github.com/inokoj/VAP-Realtime.git}).

            Since multimodal datasets include text, audio, and images simultaneously, the dataset size becomes enormous, resulting in problems with lower data transfer throughput and longer model training times. 
            To tackle this problem, we preloaded all the audio and image data into RAM at the beginning of the training and read each chunk into GPU memory (VRAM). 
            As a result, our training required about 200GB of RAM.
            Regarding the optimizer, we used the AdamW optimizer with a weight decay of 0.001 and a learning rate of 3.63e-4. 
            To stabilize the learning curve, we used the \textit{ReduceLROnPlateau} function with a reducing factor of 0.5 and a patience of 10 steps. 
            We used two Titan XP GPUs with batch sizes of 256 for training. 
            Due to GPU memory limitations, we used gradient accumulation (accumulation step: 16) to satisfy the batch size while preventing out-of-memory errors. 
            We set early stopping epochs to 5. 
            We used the same parameters as in Onishi's paper for the hyperparameters to train the Ekstedt and Onishi models. 

            We evaluated the performance of models with balanced accuracies in shift-hold prediction, short-long prediction, shift prediction, and backchannel prediction. 
            These evaluation metrics were proposed in the previous paper \cite{Ekstedt2022-a}. 
            We used the same hyperparameters for these metrics.
            Shift-hold prediction estimates whether the upcoming speech frames (frames in t+1 second) will be by the same person as the previous frames (frames in t-1 second) or a different person (hold and shift, respectively) at the silence frame with time index t in-between. 
            Short-long prediction estimates whether the current utterance lasts more than 1 second.
            We treat it as long when it equals or exceeds 1 second and is short for the others. 
            Shift prediction estimates whether a turn shift occurs within 2 seconds at the current 0.5-second prediction region. 
            Backchannel prediction estimates whether the upcoming 1-second frames are backchannel at the current 0.5-second prediction region. 
            This metric defines a backchannel as a 1-second speaking chunk surrounded by 1-second pre-silence and 2 seconds post-silence.

    \section{Result and discussion}

        Table~\ref{tab:performance} shows class-balanced accuracies of the trained models, where \textit{S/H} indicates shift-hold prediction, \textit{S/L} indicates short-long prediction, \textit{SP} indicates shift prediction, and \textit{BC} indicates backchannel prediction. 
        For SP and BC, our models (\textit{Proposed1} and \textit{Proposed3}) outperformed the baselines, where the difference lies in the use of facial embeddings from the pre-trained encoder. 
        For S/H and S/L, \textit{Baseline2} series (Baseline2\_1 and Baseline2\_2) showed the best performances for each. 
        Although our models could not beat the \textit{Baseline2} models on S/H and S/L, those results are less critical for integrating embodied conversational agent systems or robots, since we need SP and BC, which are direct estimates of when the systems should take action. 
        We observed the same tendency with the class-balanced accuracies for F1 scores. 

        In terms of the contribution of face AU and image encoder, \textit{Baseline2} models and \textit{Proposed3} model use the same face AU input; thus, the difference in score is not from the face AU but from the face image encoder. 
        While the face AU improves the SP score in general, adding the encoder further enhances it, as seen in the SP score for the \textit{Proposed3} model. 
        When comparing \textit{Baseline2\_1} and \textit{Baseline2\_2}, we can observe that the head-gaze-body features negatively impact the SP score. However, combining these features with the face encoder achieved the highest score. 
        The dimensions of the body features are comparatively larger than those of the head and gaze. 
        We can assume that the improvement stems from the interaction between the face encoder and body signals, implying that further improvement could arise with the inclusion of a body encoder.

        Regarding backchannel prediction, all models demonstrated lower accuracy compared to other metrics. 
        This might be due to the imperfect definition of the backchannel in VAP models. 
        Although the backchannel in turn-taking behavior metrics is defined solely in terms of the duration of speaking segments and the surrounding silence chunks, there are more types of backchannels in social communication. 
        For example, from the forms and functional perspective, backchannels can be classified as continuers or acknowledgment (e.g., uh-huh, yeah) or assessments or evaluations (e.g., Really? That's amazing!) \cite{Schegloff1995, Goodwin1986}. 
        Based on this classification of backchannel types, recent work predicting the continuers and assessment backchannels has shown the need for different context inputs for each kind \cite{Inoue2024-c}. 
        This implies the necessity of considering backchannel types for behavior modeling.
        Furthermore, backchannels can be multimodal. 
        In addition to verbal backchannels, nodding or facial expressions can also serve as a backchannel. 
        Therefore, we should predict not only backchannel timings, but also types of backchannels for improvements. 

        In terms of practical integration, replacing the input features of action units with facial image sequences also reduces the implementation complexity, which is one of the essential factors for practical applications. 
        Although some of the best-performing models (\textit{Baseline2\_1}, \textit{Baseline2\_2}, and \textit{Proposed3}) still rely on external software such as Openface or OpenPose, these software preparations are massive burdens in complex spoken dialogue systems. 
        These two software programs are implemented in C++, making the integration of Python-based neural networks more complex. 
        Therefore, replacing features that rely on external software with pre-trained encoders based on raw signals offers more agile architectures.
        To mitigate this aspect, our model relies solely on facial image preprocessing models (face detection and facial landmark detection), which can be easily replaced with improved models in the future. 

    \section{Conclusion}

        This paper investigated the performance of the voice activity projection model enhanced by a pretrained facial image encoder. 
        By incorporating a face image encoder, the proposed models demonstrated competitive or even superior performance compared to the multimodal state-of-the-art (SoTA) models. 
        The result reinforced the effectiveness of using pretrained facial expression encoders in turn-taking behavior predictions. 
        Based on this investigation, we aim to integrate further pre-trained encoders on other nonverbal modalities to achieve superior natural turn-taking systems. 
        Please note that there is room to improve resilience by mixing multilingual data to train models, as Inoue previously validated \cite{Inoue2024-b}.

    \section{Acknowledgements}
        
        This work was funded by the French ANR (project ENHANCER ANR22-CE17-0036-02).  
    
    \bibliographystyle{IEEEtran}
    \bibliography{interspeech2025}

\begin{thebibliography}{10}
\providecommand{\url}[1]{#1}
\csname url@samestyle\endcsname
\providecommand{\newblock}{\relax}
\providecommand{\bibinfo}[2]{#2}
\providecommand{\BIBentrySTDinterwordspacing}{\spaceskip=0pt\relax}
\providecommand{\BIBentryALTinterwordstretchfactor}{4}
\providecommand{\BIBentryALTinterwordspacing}{\spaceskip=\fontdimen2\font plus
\BIBentryALTinterwordstretchfactor\fontdimen3\font minus \fontdimen4\font\relax}
\providecommand{\BIBforeignlanguage}[2]{{%
\expandafter\ifx\csname l@#1\endcsname\relax
\typeout{** WARNING: IEEEtran.bst: No hyphenation pattern has been}%
\typeout{** loaded for the language `#1'. Using the pattern for}%
\typeout{** the default language instead.}%
\else
\language=\csname l@#1\endcsname
\fi
#2}}
\providecommand{\BIBdecl}{\relax}
\BIBdecl

\bibitem{Sacks1974}
H.~Sacks, E.~A. Schegloff, and G.~Jefferson, ``A simplest systematics for the organization of turn-taking for conversation,'' \emph{Language (Baltim.)}, vol.~50, no.~4, pp. 696--735, 1974.

\bibitem{O-Connell1990}
D.~C. O'Connell, S.~Kowal, and E.~Kaltenbacher, ``\BIBforeignlanguage{en}{Turn-taking: A critical analysis of the research tradition},'' \emph{\BIBforeignlanguage{en}{J. Psycholinguist. Res.}}, vol.~19, no.~6, pp. 345--373, Nov. 1990.

\bibitem{Duncan1972}
S.~Duncan, ``Some signals and rules for taking speaking turns in conversations,'' \emph{Journal of Personality and Social Psychology}, vol.~23, no.~2, pp. 283--292, Aug. 1972.

\bibitem{Ford1996}
C.~E. Ford and S.~A. Thompson, ``Interactional units in conversation: Syntactic, intonational, and pragmatic resources for the management of turns,'' \emph{Studies in interactional sociolinguistics}, vol.~13, pp. 134--184, 1996.

\bibitem{Ward2019}
N.~G. Ward, \emph{Prosodic patterns in English conversation}.\hskip 1em plus 0.5em minus 0.4em\relax Cambridge University Press, 2019.

\bibitem{Clark2002}
H.~H. Clark and J.~E. Fox~Tree, ``\BIBforeignlanguage{en}{Using uh and um in spontaneous speaking},'' \emph{\BIBforeignlanguage{en}{Cognition}}, vol.~84, no.~1, pp. 73--111, May 2002.

\bibitem{Ward1996}
N.~Ward, ``Using prosodic clues to decide when to produce back-channel utterances,'' in \emph{Proceeding of Fourth International Conference on Spoken Language Processing. ICSLP '96}, vol.~3.\hskip 1em plus 0.5em minus 0.4em\relax IEEE, 1996, pp. 1728--1731 vol.3.

\bibitem{Padilha2003}
E.~Padilha and J.~Carletta, ``Nonverbal behaviours improving a simulation of small group discussion,'' in \emph{The 1st Nordic Symposium on Multimodal Communication}, Sep. 2003, pp. 93--105.

\bibitem{Robins2006}
B.~Robins, K.~Dautenhahn, C.~L. Nehaniv, N.~A. Mirza, D.~Francois, and L.~Olsson, ``Sustaining interaction dynamics and engagement in dyadic child-robot interaction kinesics: lessons learnt from an exploratory study,'' in \emph{ROMAN 2005. IEEE International Workshop on Robot and Human Interactive Communication, 2005}.\hskip 1em plus 0.5em minus 0.4em\relax IEEE, 2006, pp. 716--722.

\bibitem{Cafaro2016}
A.~Cafaro, N.~Glas, and C.~Pelachaud, ``The effects of interrupting behavior on interpersonal attitude and engagement in dyadic interactions,'' in \emph{Proceedings of the 2016 International Conference on Autonomous Agents \& Multiagent Systems}, 2016, pp. 911--920.

\bibitem{Yang2023}
L.~Yang, C.~Achard, and C.~Pelachaud, ``Now or when? interruption timing prediction in dyadic interaction,'' in \emph{Proceedings of the 23rd ACM International Conference on Intelligent Virtual Agents}, 2023, pp. 1--4.

\bibitem{ter-Maat2010}
M.~ter Maat, K.~P. Truong, and D.~Heylen, ``How turn-taking strategies influence users’ impressions of an agent,'' in \emph{Intelligent Virtual Agents}, ser. Lecture notes in computer science.\hskip 1em plus 0.5em minus 0.4em\relax Berlin, Heidelberg: Springer Berlin Heidelberg, 2010, pp. 441--453.

\bibitem{Levinson2015}
S.~Levinson and F.~Torreira, ``Timing in turn-taking and its implications for processing models of language,'' \emph{Front. Psychol.}, vol.~6, Jun. 2015.

\bibitem{Meyer2023}
A.~S. Meyer, ``\BIBforeignlanguage{en}{Timing in conversation},'' \emph{\BIBforeignlanguage{en}{J. Cogn.}}, vol.~6, no.~1, p.~20, Apr. 2023.

\bibitem{DeVault2009}
D.~DeVault, K.~Sagae, and D.~Traum, ``Can {I} finish?: learning when to respond to incremental interpretation results in interactive dialogue,'' in \emph{Proceedings of the SIGDIAL 2009 Conference on The 10th Annual Meeting of the Special Interest Group on Discourse and Dialogue - SIGDIAL '09}.\hskip 1em plus 0.5em minus 0.4em\relax Morristown, NJ, USA: Association for Computational Linguistics, Sep. 2009, pp. 11--20.

\bibitem{Hough2016}
J.~Hough and D.~Schlangen, ``Investigating fluidity for human-robot interaction with real-time, real-world grounding strategies,'' in \emph{Proceedings of the 17th Annual Meeting of the Special Interest Group on Discourse and Dialogue}, R.~Fernandez, W.~Minker, G.~Carenini, R.~Higashinaka, R.~Artstein, and A.~Gainer, Eds.\hskip 1em plus 0.5em minus 0.4em\relax Association for Computational Linguistics, Sep. 2016, pp. 288--298.

\bibitem{Lala2017}
D.~Lala, P.~Milhorat, K.~Inoue, M.~Ishida, K.~Takanashi, and T.~Kawahara, ``Attentive listening system with backchanneling, response generation and flexible turn-taking,'' in \emph{Proceedings of the 18th Annual SIGdial Meeting on Discourse and Dialogue}.\hskip 1em plus 0.5em minus 0.4em\relax Stroudsburg, PA, USA: Association for Computational Linguistics, 2017.

\bibitem{Ekstedt2022-a}
E.~Ekstedt and G.~Skantze, ``Voice activity projection: Self-supervised learning of turn-taking events,'' in \emph{Interspeech 2022}.\hskip 1em plus 0.5em minus 0.4em\relax ISCA: ISCA, Sep. 2022, pp. 5190--5194.

\bibitem{Ishii2013}
R.~Ishii, K.~Otsuka, S.~Kumano, M.~Matsuda, and J.~Yamato, ``Predicting next speaker and timing from gaze transition patterns in multi-party meetings,'' in \emph{Proceedings of the 15th ACM on International conference on multimodal interaction}.\hskip 1em plus 0.5em minus 0.4em\relax New York, NY, USA: ACM, Dec. 2013.

\bibitem{Tian2018}
L.~Tian, Q.~Jia, and Z.~Zhu, ``Predicting turn-taking by compact gazing transition patterns in multiparty conversation,'' in \emph{Image and Video Technology}, ser. Lecture notes in computer science.\hskip 1em plus 0.5em minus 0.4em\relax Cham: Springer International Publishing, 2018, pp. 437--447.

\bibitem{Ishii2019}
R.~Ishii, K.~Otsuka, S.~Kumano, R.~Higashinaka, and J.~Tomita, ``\BIBforeignlanguage{en}{Prediction of who will be next speaker and when using mouth-opening pattern in multi-party conversation},'' \emph{\BIBforeignlanguage{en}{Multimodal Technol. Interact.}}, vol.~3, no.~4, p.~70, Oct. 2019.

\bibitem{Onishi2023}
K.~Onishi, H.~Tanaka, and S.~Nakamura, ``Multimodal voice activity prediction: Turn-taking events detection in expert-novice conversation,'' in \emph{International Conference on Human-Agent Interaction}.\hskip 1em plus 0.5em minus 0.4em\relax New York, NY, USA: ACM, Dec. 2023, pp. 13--21.

\bibitem{Wang2024}
J.~Wang, L.~Chen, A.~Khare, A.~Raju, P.~Dheram, D.~He, M.~Wu, A.~Stolcke, and V.~Ravichandran, ``Turn-taking and backchannel prediction with acoustic and large language model fusion,'' \emph{arXiv [cs.CL]}, Jan. 2024.

\bibitem{Pinto2024}
M.~J. Pinto and T.~Belpaeme, ``Predictive turn-taking: Leveraging language models to anticipate turn transitions in human-robot dialogue,'' in \emph{2024 33rd IEEE International Conference on Robot and Human Interactive Communication (ROMAN)}.\hskip 1em plus 0.5em minus 0.4em\relax IEEE, Aug. 2024, pp. 1733--1738.

\bibitem{Xie2024}
Z.~Xie and C.~Wu, ``Mini-omni: Language models can hear, talk while thinking in streaming,'' \emph{arXiv [cs.AI]}, Aug. 2024.

\bibitem{Defossez2024}
A.~Défossez, L.~Mazaré, M.~Orsini, A.~Royer, P.~Pérez, H.~Jégou, E.~Grave, and N.~Zeghidour, ``Moshi: a speech-text foundation model for real-time dialogue,'' \emph{arXiv [eess.AS]}, Sep. 2024.

\bibitem{GeminiLive}
\BIBentryALTinterwordspacing
Google. [Online]. Available: \url{https://support.google.com/gemini/answer/15274899?hl=en&co=GENIE.Platform%3DAndroid}
\BIBentrySTDinterwordspacing

\bibitem{ChatGPTVoiceChat}
\BIBentryALTinterwordspacing
OpenAI. [Online]. Available: \url{https://help.openai.com/en/articles/8400625-voice-mode-faq}
\BIBentrySTDinterwordspacing

\bibitem{Inoue2024-a}
K.~Inoue, B.~Jiang, E.~Ekstedt, T.~Kawahara, and G.~Skantze, ``Real-time and continuous turn-taking prediction using voice activity projection,'' \emph{arXiv [cs.CL]}, Jan. 2024.

\bibitem{Inoue2024-b}
------, ``Multilingual turn-taking prediction using voice activity projection,'' \emph{arXiv [cs.CL]}, Mar. 2024.

\bibitem{Ekstedt2022-b}
E.~Ekstedt and G.~Skantze, ``How much does prosody help turn-taking? investigations using voice activity projection models,'' \emph{arXiv [eess.AS]}, Sep. 2022.

\bibitem{Baltrusaitis2016}
T.~Baltrusaitis, P.~Robinson, and L.-P. Morency, ``{OpenFace}: An open source facial behavior analysis toolkit,'' in \emph{2016 IEEE Winter Conference on Applications of Computer Vision (WACV)}.\hskip 1em plus 0.5em minus 0.4em\relax IEEE, Mar. 2016, pp. 1--10.

\bibitem{Cao2021}
Z.~Cao, G.~Hidalgo, T.~Simon, S.-E. Wei, and Y.~Sheikh, ``Openpose: Realtime multi-person 2d pose estimation using part affinity fields,'' \emph{IEEE Transactions on Pattern Analysis and Machine Intelligence}, vol.~43, no.~1, pp. 172--186, 2021.

\bibitem{Zhao2021}
Z.~Zhao and Q.~Liu, ``Former-dfer: Dynamic facial expression recognition transformer,'' in \emph{Proceedings of the 29th ACM International Conference on Multimedia}, 2021, pp. 1553--1561.

\bibitem{Riviere2020}
M.~Rivière, A.~Joulin, P.-E. Mazaré, and E.~Dupoux, ``Unsupervised pretraining transfers well across languages,'' in \emph{ICASSP 2020 - 2020 IEEE International Conference on Acoustics, Speech and Signal Processing (ICASSP)}, 2020, pp. 7414--7418.

\bibitem{King2009}
D.~E. King, ``Dlib-ml: A machine learning toolkit,'' \emph{J. Mach. Learn. Res.}, vol.~10, no.~60, pp. 1755--1758, Dec. 2009.

\bibitem{Cafaro2017}
A.~Cafaro, J.~Wagner, T.~Baur, S.~Dermouche, M.~Torres~Torres, C.~Pelachaud, E.~André, and M.~Valstar, ``The {NoXi} database: multimodal recordings of mediated novice-expert interactions,'' in \emph{Proceedings of the 19th ACM International Conference on Multimodal Interaction}.\hskip 1em plus 0.5em minus 0.4em\relax New York, NY, USA: ACM, Nov. 2017.

\bibitem{pytorch_lightning}
\BIBentryALTinterwordspacing
W.~Falcon and T.~P.~L. team, ``{PyTorch Lightning},'' 12 2019. [Online]. Available: \url{https://github.com/Lightning-AI/pytorch-lightning}
\BIBentrySTDinterwordspacing

\bibitem{Schegloff1995}
\BIBentryALTinterwordspacing
E.~A. Schegloff, ``Discourse as an interactional achievement iii: The omnirelevance of action,'' \emph{Research on Language and Social Interaction}, vol.~28, no.~3, pp. 185--211, 1995. [Online]. Available: \url{https://doi.org/10.1207/s15327973rlsi2803_2}
\BIBentrySTDinterwordspacing

\bibitem{Goodwin1986}
C.~Goodwin, ``\BIBforeignlanguage{en}{Between and within: Alternative sequential treatments of continuers and assessments},'' \emph{\BIBforeignlanguage{en}{Hum. Stud.}}, vol.~9, no. 2-3, pp. 205--217, 1986.

\bibitem{Inoue2024-c}
K.~Inoue, D.~Lala, G.~Skantze, and T.~Kawahara, ``Yeah, un, oh: Continuous and real-time backchannel prediction with fine-tuning of voice activity projection,'' \emph{arXiv [cs.CL]}, Oct. 2024.

\end{thebibliography}

\end{document}